\documentstyle[aclap, qtree]{article}

\title{A Linear Observed Time Statistical Parser Based on Maximum Entropy Models}

\author{Adwait Ratnaparkhi\thanks{\ The author acknowledges the support of ARPA grant N66001-94C-6043.}\\
Dept.~of Computer and Information Science\\
University of Pennsylvania\\
200 South 33rd Street\\
Philadelphia, PA 19104-6389\\
{\tt adwait@unagi.cis.upenn.edu}}

\begin{document}

\bibliographystyle{fullname}

\maketitle
\begin{abstract}
This paper presents a statistical parser for 
natural language that obtains a parsing accuracy---roughly 
87\% precision and 86\% recall---which surpasses
the best previously published results on
the Wall St.~Journal domain.
The parser itself requires very little human intervention,
since the information it uses to make parsing decisions
is specified in a concise and simple manner, and is combined
in a fully automatic way under the maximum entropy framework.
The observed running time of the parser on a test sentence 
is linear with respect to the sentence length.
Furthermore, the parser returns several scored
parses for a sentence, and this paper shows that
a scheme to pick the best parse from the 20
highest scoring parses {\em could} yield a 
dramatically higher accuracy of 93\% precision and recall.
\end{abstract}

\section{Introduction}
\newcommand{\proc}[1]{{\sc #1}}
\newcommand{\deriv}[1]{{\tt #1}}
\newcommand{\feat}[2]{{\sf #1}($#2$)}
\newcommand{\Feat}[1]{{\sf #1}}

This paper presents a statistical parser for natural language that finds one or
more scored syntactic parse trees for a given input sentence.
The parsing accuracy---roughly 87\% precision and 86\% recall---surpasses
the best previously published results on
the Wall St.~Journal domain.
The parser consists of the following
three conceptually distinct parts:
\begin{enumerate}
\item A set of procedures that use certain actions to incrementally construct parse trees.

\item A set of maximum entropy models that compute
probabilities of the above actions, and effectively ``score'' parse trees.

\item A search heuristic which attempts to find
the highest scoring parse tree for a given input
sentence.
\end{enumerate}
The maximum entropy models used here are similar in form to those in 
\cite{maxent:tagger,maxent:mt,maxent:lm}.
The models compute the probabilities of actions
based on certain syntactic characteristics, or {\em features},
of the current context. 
The features used here are defined in a concise and simple manner,
and their relative importance is determined automatically
by applying a training procedure on a corpus of syntactically
annotated sentences, such as the Penn Treebank~\cite{treebank}.
Although creating the annotated corpus requires much linguistic expertise,
creating the feature set for the parser itself requires very little linguistic effort.

Also, the search heuristic is very simple, and
its observed running time on a test sentence 
is linear with respect to the sentence length.
Furthermore, the search heuristic returns several scored
parses for a sentence, and this paper shows that
a scheme to pick the best parse from the 20
highest scoring parses {\em could} yield a 
dramatically higher accuracy of 93\% precision and recall.

Sections \ref{tbproc}, \ref{maxentsection}, and \ref{searchsection}
describe the tree-building procedures, the maximum entropy models,
and the search heuristic, respectively. Section \ref{expsection}
describes experiments with the Penn Treebank and section \ref{discussion}
compares this paper with previously published works.

\begin{table*}
\begin{center}
\begin{tabular}{|c|c|p{2in}|p{2in}|} \hline
Pass&Procedure&Actions&Description\\ \hline
First Pass&\proc{tag}&A POS tag in tag set&Assign POS Tag to
word\\ \hline
Second Pass&\proc{chunk}&{\tt Start X}, {\tt Join X}, {\tt Other}&Assign 
Chunk tag to POS tag and word\\ \hline
Third Pass&\proc{build}&{\tt Start X}, {\tt Join X}, where {\tt X} is a
constituent label in label set&Assign current tree to start a new constituent,
or to join the previous one\\ \cline{2-4}
&           \proc{check}&{\tt Yes}, {\tt No}&Decide if current constituent is
complete\\ \hline
\end{tabular}
\caption{Tree-Building Procedures of Parser}
\label{Procedures}
\end{center}
\end{table*}

\section{Procedures for Building Trees}
\label{tbproc}

The parser uses four procedures, 
\proc{tag}, \proc{chunk}, \proc{build}, and \proc{check}, 
that incrementally build parse trees with their actions.
The procedures are applied in three left-to-right passes over 
the input sentence; the first pass applies \proc{tag}, the
second pass applies \proc{chunk}, and the third pass applies
\proc{build} and \proc{check}.
The passes, the procedures they apply, and the actions of the
procedures are summarized in table~\ref{Procedures} and described below.

The actions of the procedures are designed so that any possible complete parse 
tree $T$ for the input sentence corresponds to {\em exactly one} sequence of actions; call this 
sequence the {\em derivation} of $T$.
Each procedure, when given a derivation $d = \{ a_1 \dots a_n \}$, predicts
some action $a_{n+1}$ to create a new derivation $d' = \{ a_1 \dots a_{n+1} \}$.
Typically, the procedures postulate many different values for $a_{n+1}$, 
which cause the parser to explore many different derivations when parsing an input
sentence.
But for demonstration purposes, 
figures~\ref{initforest}--\ref{checkforest} trace one possible derivation
for the sentence ``I saw the man with the telescope'', using the
part-of-speech (POS) tag set and constituent label set of the Penn treebank.

\subsection{First Pass}

\begin{figure*}
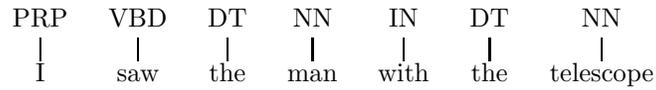

\begin{center}
\qtreecenterfalse
\Tree [.PRP I ]
\Tree [.VBD saw ]
\Tree [.DT the ]	
\Tree [.NN man ]	
\Tree [.IN with ]
\Tree [.DT the ]
\Tree [.NN telescope ]
\end{center}
\caption{The result after First Pass}
\label{forestafterfirst}
\end{figure*}

The first pass takes an input sentence, shown in figure~\ref{initforest}, 
and uses \proc{tag} to assign each word a POS tag.
The result of applying \proc{tag} to each word is shown in figure~\ref{forestafterfirst}.

\subsection{Second Pass}

\begin{figure*}
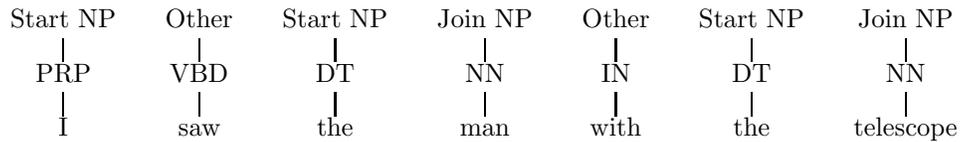

\begin{center}
\qtreecenterfalse
\Tree [.{Start NP} [.PRP I ] ]
\Tree [.{Other} [.VBD saw ] ]
\Tree [.{Start NP} [.DT the ] ]
\Tree [.{Join NP} [.NN man ] ]
\Tree [.{Other} [.IN with ] ]
\Tree [.{Start NP} [.DT the ] ]
\Tree [.{Join NP} [.NN telescope ] ]
\end{center}
\caption{The result after Second Pass}
\label{forestaftersecond}
\end{figure*}

\begin{figure*}
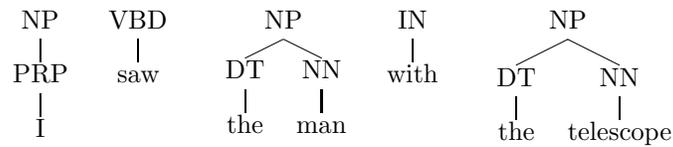

\begin{center}
\qtreecenterfalse
\Tree [.NP [.PRP I ] ]
\Tree [.VBD saw ]
\Tree [.NP [.DT the ] [.NN man ] ]
\Tree [.IN with ]
\Tree [.NP [.DT the ] [.NN telescope ] ]
\end{center}
\caption{The result of chunk detection}
\label{forestaftersimpleNP}
\end{figure*}

The second pass takes the output of the first pass and 
uses \proc{chunk} to determine the ``flat'' phrase chunks of the sentence, where a 
phrase is ``flat'' if and only if it is a constituent
whose children consist solely of POS tags.
Starting from the left, \proc{chunk} assigns
each (word,POS tag) pair a ``chunk'' tag, either
{\tt Start X}, {\tt Join X}, or {\tt Other}. 
Figure~\ref{forestaftersecond} shows the result after
the second pass.
The chunk tags are then used for chunk detection, in which
any consecutive sequence of words $w_m \dots w_n$ ($m \leq n$) 
are grouped into a ``flat'' chunk $X$ if $w_m$ has been assigned {\tt Start X} and $w_{m+1} \dots w_n$ 
have all been assigned {\tt Join X}. The result of chunk detection,
shown in figure~\ref{forestaftersimpleNP},
is a forest of trees and serves as the input to the third pass.

\begin{table}
\begin{center}
\begin{tabular}{|c|c|p{1in}|} \hline
Procedure&Actions&Similar Shift-Reduce Parser Action\\ \hline
\proc{check}&{\tt No}&shift\\ \hline
\proc{check}&{\tt Yes}&reduce $\alpha$, where $\alpha$ is CFG rule of proposed constituent\\ \hline
\proc{build}&{\tt Start X}, {\tt Join X}&Determines $\alpha$ for subsequent
reduce operations\\ \hline
\end{tabular}
\caption{Comparison of \proc{build} and \proc{check} to operations of a shift-reduce parser}
\label{shiftreduce}
\end{center}
\end{table}

\begin{figure*}
\begin{center}
I saw the man with the telescope 
\end{center}
\caption{Initial Sentence}
\label{initforest}
\end{figure*}

\begin{figure*}
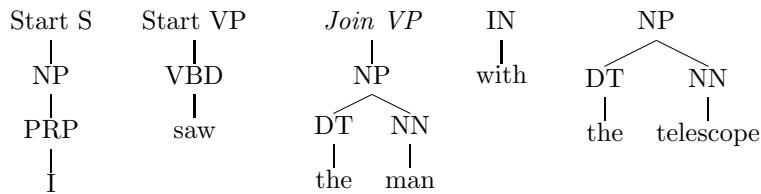

\begin{center}
\qtreecenterfalse
\Tree [.{Start S} [.NP [.PRP I ]]]
\Tree [.{Start VP}  [.VBD saw ] ]
\Tree [.{\em Join VP} [.NP [.DT the ] [.NN man ] ]]
\Tree [.IN with ]
\Tree [.NP [.DT the ] [.NN telescope ] ]
\end{center}
\caption{An application of \proc{build} in which {\tt Join VP} is the action}
\label{buildforest}
\end{figure*}

\begin{figure*}
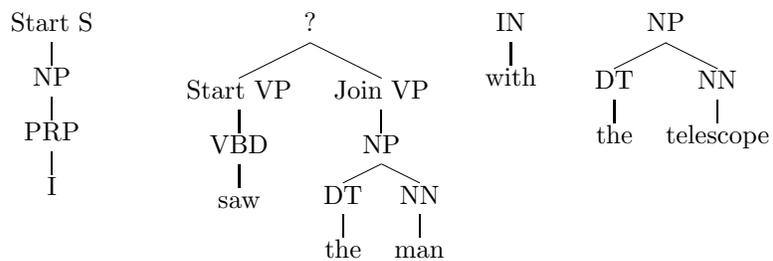

\begin{center}
\qtreecenterfalse
\Tree [.{Start S} [.NP [.PRP I ]]]
\Tree [.{?} [.{Start VP}  [.VBD saw ] ] [.{Join VP} [.NP [.DT the ] [.NN man ] ]]]
\Tree [.IN with ]
\Tree [.NP [.DT the ] [.NN telescope ] ]
\end{center}
\caption{The most recently proposed constituent (shown under ?)}
\label{proposedcons}
\end{figure*}

\begin{figure*}
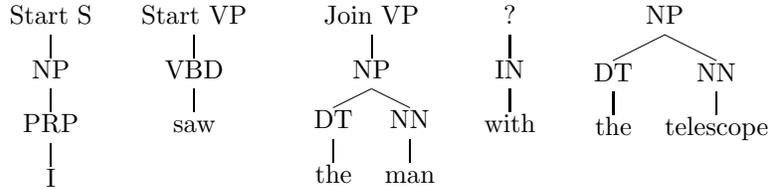

\begin{center}
\qtreecenterfalse
\Tree [.{Start S} [.NP [.PRP I ]]]
\Tree [.{Start VP}  [.VBD saw ] ]
\Tree [.{Join VP} [.NP [.DT the ] [.NN man ] ]]
\Tree [.{?} [.IN with ] ]
\Tree [.NP [.DT the ] [.NN telescope ] ]
\end{center}
\caption{An application of \proc{check} in which {\tt No} is the action,
indicating that the proposed constituent in figure~\ref{proposedcons}
is {\em not} complete. \proc{build} will now process the tree marked with ?}
\label{checkforest}
\end{figure*}

\subsection{Third Pass}

The third pass always alternates between the use of
\proc{build} and \proc{check}, and
completes any remaining constituent structure. 
\proc{build} 
decides whether a tree will start a new constituent or join the 
incomplete constituent immediately to its left.
Accordingly, it annotates the tree with either {\tt Start X}, where {\tt X} is any
constituent label, or with {\tt Join X}, where {\tt X}
matches the label of the incomplete constituent to the left.
\proc{build} always processes the leftmost tree without any
{\tt Start X} or {\tt Join X} annotation.
Figure~\ref{buildforest} shows an application of
\proc{build} in which the action is {\tt Join VP}.
After \proc{build}, control passes to \proc{check},
which finds the most recently proposed constituent,
and decides if it is complete.
The most recently proposed constituent, shown in
figure~\ref{proposedcons},  is 
the rightmost sequence of trees $t_m \dots t_n$ ($m \leq n$)
such that $t_m$ is annotated with {\tt Start X} and
$t_{m+1}\dots t_n$ are annotated with {\tt Join X}.
If \proc{check} decides {\tt yes}, then the proposed constituent
takes its place in the forest as an actual constituent, on which
\proc{build} does its work. Otherwise, the constituent is not finished and \proc{build} processes 
the next tree in the forest, $t_{n+1}$.
\proc{check} always answers {\tt no} if the proposed constituent
is a ``flat'' chunk, since such constituents must be formed in 
the second pass.
Figure~\ref{checkforest} shows the result when
\proc{check} looks at the proposed constituent in figure~\ref{proposedcons} 
and decides {\tt No}.
The third pass terminates when \proc{check} is presented
a constituent that spans the entire sentence.

Table~\ref{shiftreduce} compares the actions of \proc{build} 
and \proc{check} to the operations of a standard
shift-reduce parser.
The {\tt No} and {\tt Yes} actions of \proc{check} correspond to the
shift and reduce actions, respectively.
The important difference is that while a shift-reduce parser creates a
constituent in one step (reduce $\alpha$), 
the procedures \proc{build} and \proc{check} create it over several steps in 
smaller increments.

\section{Probability Model}
\label{maxentsection}

This paper takes a ``history-based'' approach
\cite{black:hbg} where 
each tree-building procedure uses
a probability model $p(a|b)$, derived from $p(a,b)$, to weight any 
action $a$ based on the available context, or history, $b$.
First, we present a few simple categories of contextual predicates that capture
any information in $b$ that is useful for predicting $a$. 
Next, the predicates are used to extract a set of features from
a corpus of manually parsed sentences.
Finally, those features are combined under the maximum entropy framework,
yielding $p(a,b)$.

\subsection{Contextual Predicates}
\label{featuresection}

Contextual predicates are functions that check
for the presence or absence of useful information 
in a context $b$ and return true or false accordingly.
The comprehensive guidelines, or templates, for the contextual predicates
of each tree building procedure are given in table~\ref{featuretable}.
The templates use
indices relative to the tree that is currently being modified.
For example, if the current tree is the $5$th tree, 
\feat{cons}{-2} looks at the constituent label,
head word, and start/join annotation 
of the $3$rd tree in the forest.
The actual contextual predicates are generated automatically
by scanning the derivations of the trees in the manually parsed 
corpus with the templates.
For example, an actual contextual predicate
based on the template \feat{cons}{0} might be ``Does \feat{cons}{0} = \{ {\tt NP, he} \} ?''
Constituent head words are found, when necessary, with the 
algorithm in \cite{magerman:acl95}.

Contextual predicates which look at head words, or especially pairs of head words,
may not be reliable predictors for the procedure actions
due to their sparseness in the training sample.
Therefore, for each lexically based contextual predicate,
there also exist one or more corresponding less specific, or ``backed-off'',
contextual predicates which look at the same context, but
{\em omit} one or more words.
For example, the contexts \feat{cons}{0, 1^*}, \feat{cons}{0^*, 1}, \feat{cons}{0^*,1^*}
are the same as \feat{cons}{0,1} but omit references 
to the head word of the 1st tree, the 0th tree, and both
the 0th and 1st tree, respectively.
The backed-off contextual predicates should allow the model to provide
reliable probability estimates when the words in the history are rare.
Backed-off predicates are not enumerated in table~\ref{featuretable},
but their existence is indicated with a $^*$ and $^\dagger$.

\begin{table*}
{
\begin{tabular}{|c|c|p{2.5in}|p{1.75in}|} \hline
Model&Categories&Description&Templates Used\\ \hline
\proc{tag}&\multicolumn{3}{c|}{See \cite{maxent:tagger}}\\ \hline
\proc{chunk}&\feat{chunkandpostag}{n}$^*$&The word, POS tag, and chunk tag of
$n$th leaf. Chunk tag omitted if $n \geq 0$.&
\feat{chunkandpostag}{0}, \feat{chunkandpostag}{-1}, \feat{chunkandpostag}{-2}
\feat{chunkandpostag}{1}, \feat{chunkandpostag}{2}\\ \cline{2-4}
&\feat{chunkandpostag}{m,n}$^*$&\feat{chunkandpostag}{m} \&
\feat{chunkandpostag}{n}&\feat{chunkandpostag}{-1, 0}, \feat{chunkandpostag}{0, 1} \\ \hline
\proc{build}&\feat{cons}{n}&
The head word, constituent (or POS) label, and start/join annotation of the
$n$th tree. Start/join annotation omitted if $n \geq 0$.
&\feat{cons}{0}, \feat{cons}{-1}, \feat{cons}{-2},
\feat{cons}{1}, \feat{cons}{2}\\ \cline{2-4}
&\feat{cons}{m,n}$^*$&\feat{cons}{m} \& \feat{cons}{n}&
\feat{cons}{-1, 0}, \feat{cons}{0, 1}\\ \cline{2-4}
&\feat{cons}{m, n, p}$^\dagger$&\feat{cons}{m}, \feat{cons}{n}, \&
\feat{cons}{p}.&
\feat{cons}{0, -1, -2}, \feat{cons}{0, 1, 2}, \feat{cons}{-1, 0, 1}\\ \cline{2-4}
&\Feat{punctuation}&
The constituent we could join
(1) contains a ``[''  and the current tree is a ``]'';
(2) contains a ``,'' and the current tree is a ``,'';
(3) spans the entire sentence and current tree is ``.''
&\Feat{bracketsmatch}, \Feat{iscomma}, \Feat{endofsentence}\\ \hline
\proc{check}&\feat{checkcons}{n}$^*$&
The head word, constituent (or POS) label of the $n$th tree,
and the label of proposed constituent.
$begin$ and $last$ are first and last child (resp.) of proposed constituent.&
\feat{checkcons}{last}, \feat{checkcons}{begin}\\ \cline{2-4}
&\feat{checkcons}{m,n}$^*$&\feat{checkcons}{m} \& \feat{checkcons}{n}&
\feat{checkcons}{i, last}, $begin \leq i < last$\\ \cline{2-4}
&\Feat{production}&Constituent label of parent ($X$), and constituent or POS
labels of children ($X_1 \dots X_n$) of proposed constituent
&\Feat{production=$X \rightarrow X_1 \dots X_n$}\\ \cline{2-4}
&\feat{surround}{n}$^*$&POS tag and word of the $n$th leaf to the
left of the constituent, if $n < 0$, or to the right of
the constituent, if $n > 0$&\feat{surround}{1}, \feat{surround}{2},
\feat{surround}{-1}, \feat{surround}{-2} \\ \hline
\end{tabular}
}
\caption{Contextual Information Used by Probability Models 
($^*$ = all backed-off contexts are used, $^\dagger$ = only backed-off
contexts that include head word of current tree, i.e., 0th tree, are used)}
\label{featuretable}
\end{table*}

\subsection{Maximum Entropy Framework}

\newcommand{\mathcal}{\cal}
\newcommand{\mathsf}{\sf}
\newcommand{\procb}[1]{{\rm #1}}
\newcommand{\trsamp}[1]{{\mathcal T}_{\rm #1}}

The contextual predicates derived from the templates of table~\ref{featuretable} are used
to create the features necessary for the maximum entropy models.
The predicates for \proc{tag}, \proc{chunk}, \proc{build},
and \proc{check} are used to scan the derivations 
of the trees in the corpus to form the training
samples $\trsamp{tag}$, $\trsamp{chunk}$, $\trsamp{build}$, and
$\trsamp{check}$, respectively. Each training sample has the
form ${\mathcal T} = \{ (a_1, b_1), (a_2, b_2), \dots, (a_N, b_N) \}$,
where $a_i$ is an action of the corresponding procedure 
and $b_i$ is the list of contextual predicates that were {\tt true} in the 
context in which $a_i$ was decided. 

The training samples are respectively used to create 
the models $p_{\procb{tag}}$, $p_{\procb{chunk}}$, $p_{\procb{build}}$, and
$p_{\procb{check}}$, all of which have the form:
\begin{equation}
p(a,b) = \pi \prod_{j=1}^k \alpha_j^{f_j(a,b)} \label{form}
\end{equation}
where $a$ is some action, $b$ is some context, 
$\pi$ is a normalization constant,
$\alpha_j$ are the model parameters, $0 < \alpha_j < \infty$,
and $f_j(a,b) \in \{ 0, 1 \}$ are called {\em features}, $j = \{ 1 \dots k \}$. 
Features encode an action $a'$ as well as some contextual predicate $cp$ 
that a tree-building procedure would find useful for predicting the action $a'$. 
Any contextual predicate $cp$ derived from table~\ref{featuretable} which occurs 5 or more times in
a training sample with a particular action $a'$ is 
used to construct a feature $f_j$:
\[ f_j(a,b) = 
\left\{ 
\begin{array}{ll}
1&\mbox{if $cp(b) =$ {\tt true} \&\& $a=a'$}\\
0&\mbox{otherwise} \\
\end{array}
\right. \]
for use in the corresponding model.
Each feature $f_j$ corresponds to a parameter $\alpha_j$, which
can be viewed as a ``weight'' that reflects the importance
of the feature. 

The parameters $\{ \alpha_1 \dots \alpha_n \}$ are found automatically
with {\em Generalized Iterative Scaling}~\cite{scaling}, or GIS.
The GIS procedure, as well as the
maximum entropy and maximum likelihood properties 
of the distribution of form~(\ref{form}), are
described in detail in \cite{maxent:tech}.
In general, the maximum entropy framework puts no limitations on the kinds
of features in the model; no special estimation technique 
is required to combine features that encode 
different kinds of contextual predicates, 
like \Feat{punctuation} and \feat{cons}{0,1,2}.
As a result, experimenters need only worry about {\em what}
features to use, and not {\em how} to use them.
 
We then use the models $p_{\procb{tag}}$, $p_{\procb{chunk}}$, $p_{\procb{build}}$, and
$p_{\procb{check}}$ to define a function ${\sf score}$, which
the search procedure uses to rank derivations of incomplete and complete
parse trees.
For each model, the corresponding conditional probability is defined as usual:
\[ p(a|b) = {p(a,b) \over {\sum_{a' \in A} p(a',b)}} \]
For notational convenience, define $q$ as follows
\[ q(a|b) = 
\left\{ 
\begin{array}{ll}
p_{\procb{tag}}(a|b)&\mbox{if $a$ is an action from \proc{tag}}\\
p_{\procb{chunk}}(a|b)&\mbox{if $a$ is an action from \proc{chunk}}\\
p_{\procb{build}}(a|b)&\mbox{if $a$ is an action from \proc{build}}\\
p_{\procb{check}}(a|b)&\mbox{if $a$ is an action from \proc{check}}\\
\end{array}
\right. \]
Let ${\mathsf deriv}(T) = \{ a_1, \dots, a_n \}$ be the
derivation of a parse $T$, where $T$ is not necessarily complete, and 
where each $a_i$ is an action of some tree-building procedure.
By design, the tree-building procedures guarantee that $\{ a_1, \dots, a_n \}$
is the only derivation for the parse $T$.
Then the score of $T$ is merely the product of 
the conditional probabilities of the individual actions in
its derivation:
\[ {\mathsf score}(T) = \prod_{a_i \in {\mathsf deriv}(T)} q(a_i | b_i) \]
where $b_i$ is the context in which $a_i$ was decided. 

\section{Search}
\label{searchsection}
The search heuristic attempts to find the best parse $T^*$, defined as:
\[  T^* = \arg\max_{T \in {\mathsf trees}(S)} {\mathsf score}(T) \]
where ${\mathsf trees}(S)$ are all the complete parses for an input
sentence $S$.  

The heuristic employs a breadth-first search (BFS) which does not
explore the entire frontier, but rather, 
explores only at most the top $K$ scoring incomplete parses in the frontier,
and terminates when it has found $M$ complete parses, or when all the hypotheses
have been exhausted. 
Furthermore, if $\{a_1 \dots a_n \}$ are the possible actions
for a given procedure on a derivation with context $b$, 
and they are sorted in decreasing order according to $q(a_i | b)$,
we only consider exploring those actions $\{a_1 \dots a_m \}$
that hold most of the probability mass, where $m$ is defined
as follows:
\[ m = \max_m \sum_{i=1}^m q(a_i|b) < Q \]
and where $Q$ is a threshold less than 1. 
The search also uses a {\em Tag Dictionary} constructed from training data, 
described in \cite{maxent:tagger},
that reduces the number of actions explored by the tagging model.
Thus there are three parameters for the search heuristic, namely
$K$,$M$, and $Q$ and all experiments reported in this paper 
use $K=20$, $M=20$, and $Q=.95$\footnote{The parameters $K$,$M$, and $Q$
were optimized on ``held out'' data separate from the training and
test sets.}
Table~\ref{searchtable} describes the top $K$
BFS and the semantics of the supporting functions. 

It should be emphasized that if $K>1$, the parser does not commit to a single POS or
chunk assignment for the input sentence before building constituent structure.
All three of the passes described in
section~\ref{tbproc} are integrated in the search, i.e., when parsing
a test sentence, the input to the second pass consists of $K$ of
the best distinct POS tag assignments for the input sentence. Likewise, 
the input to the third pass consists of $K$ of the best distinct
chunk and POS tag assignments for the input sentence.
 
\begin{table*}
\begin{center}
{\tt
\begin{tabular}{lll}
advance:&$d \times V \longrightarrow d_1 \dots d_m$&/* Applies relevant tree building
procedure to $d$\\
&&and returns list of new derivations whose action\\
&&probabilities pass the threshold $Q$ */\\
insert:&$d \times h \longrightarrow$ void&/* inserts $d$ in heap $h$ */\\
extract:&$h \longrightarrow d$&/* removes and returns derivation in $h$\\
&&with highest score */\\
completed:&$d \longrightarrow$ \{true,false\}&/* returns true if and only if\\
&&$d$ is a complete derivation */ \\
\end{tabular}
\begin{tabbing}
$M = 20$\\
$K = 20$\\
$Q = .95$\\
$C =$ <empty heap>\ \ \ \ \ \ \ \ \ \ /* Heap of completed parses */\\
$h_0 = $<input sentence>\ \ \ \ \ \ \ /* $h_i$ contains derivations of length $i$ */ \\
while 	\=( $|C| < M$ )\\
	\>if \=( $\forall i,  h_i$ is empty )\\
	\>   \>then break\\	
	\>$i = \max \{i\ |\ \mbox{$h_i$ is non-empty}\}$\\
	\>$sz = \min(K, |h_i|)$\\
	\>for \=$j=1$ to $sz$\\
	\>	\>$d_1 \dots d_p$ = advance( extract($h_i$), $V$ )\\
	\> 	\>for \=$q=1$ to $p$\\
	\>	\> 	\>if \=(completed($d_q$))\\
	\>	\>	\>   \>then insert($d_q$, $C$)\\	
	\>	\>	\>   \>else insert($d_q$, $h_{i+1}$)\\
\end{tabbing}
}
\caption{Top $K$ BFS Search Heuristic}
\label{searchtable}
\end{center}
\end{table*}

\begin{figure*}
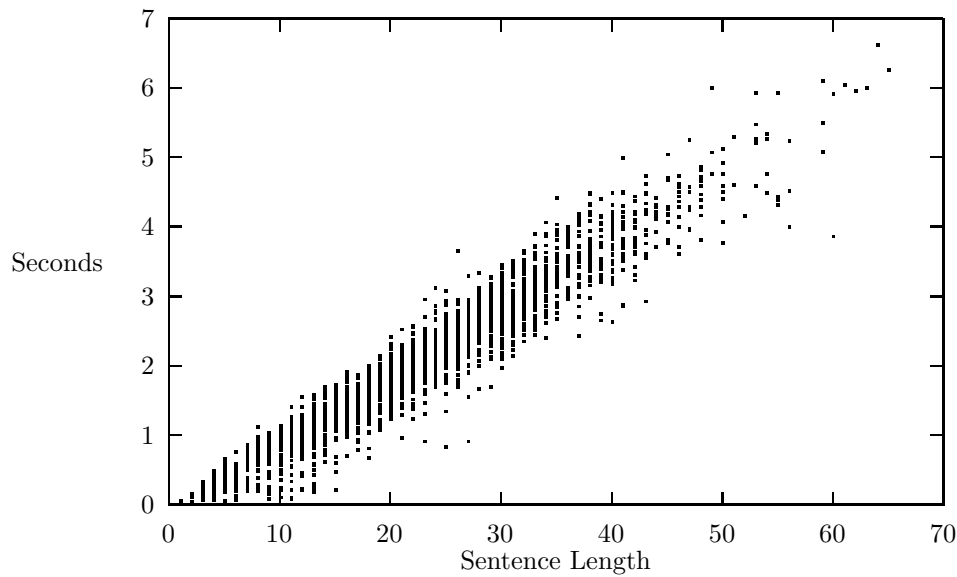

\begin{center}
\include{runningtime}
\caption{Observed running time of top $K$ BFS on Section 23 of Penn Treebank
WSJ, using one 167Mhz UltraSPARC processor and 256MB RAM of a Sun Ultra Enterprise 4000.}
\label{runningtime}
\end{center}
\end{figure*}
    
The top $K$ BFS described above exploits
the observed property that the individual 
steps of correct derivations tend to have high probabilities,
and thus avoids searching a large fraction of the search space.
Since, in practice, it only does a constant amount of work to
advance each step in a derivation, and since derivation lengths
are roughly proportional to the sentence length, we would
expect it to run in linear observed time with respect to sentence length.
Figure~\ref{runningtime} confirms our assumptions about
the linear observed running time.

\section{Experiments}
\label{expsection}

The maximum entropy parser was trained on sections 2 through 21 (roughly 40000
sentences)
of the Penn Treebank Wall St.~Journal corpus, release 2~\cite{treebank}, 
and tested on section 23 (2416 sentences) for comparison with other work. 
All trees were stripped of their semantic tags (e.g., {\tt -LOC}, {\tt
-BNF}, etc.), coreference information(e.g., {\tt *-1}), and quotation marks ({\tt ``} and {\tt ''}) for
both training and testing.
The PARSEVAL~\cite{parseval} measures compare a proposed parse $P$ with the
corresponding correct treebank parse $T$ as follows:
\begin{eqnarray*}
\mbox{Recall} &=& {\mbox{\# correct constituents in }P \over 
\mbox{\# constituents in }T}\\
\mbox{Precision} &=& {\mbox{\# correct constituents in }P \over 
\mbox{\# constituents in }P}\\
\end{eqnarray*}
A constituent in $P$ is ``correct'' if there exists a
constituent in $T$ of the same label that spans the same words.
Table~\ref{results} shows results using the PARSEVAL measures, 
as well as results using the slightly more forgiving measures of \cite{collins:acl96} and
\cite{magerman:acl95}.
Table~\ref{results} shows that the maximum entropy parser 
performs better than the parsers presented in 
\cite{collins:acl96} and \cite{magerman:acl95}\footnote{Results for SPATTER on
section 23 are reported in \cite{collins:acl96}}, 
which have the best previously published 
parsing accuracies on the Wall St. Journal domain.

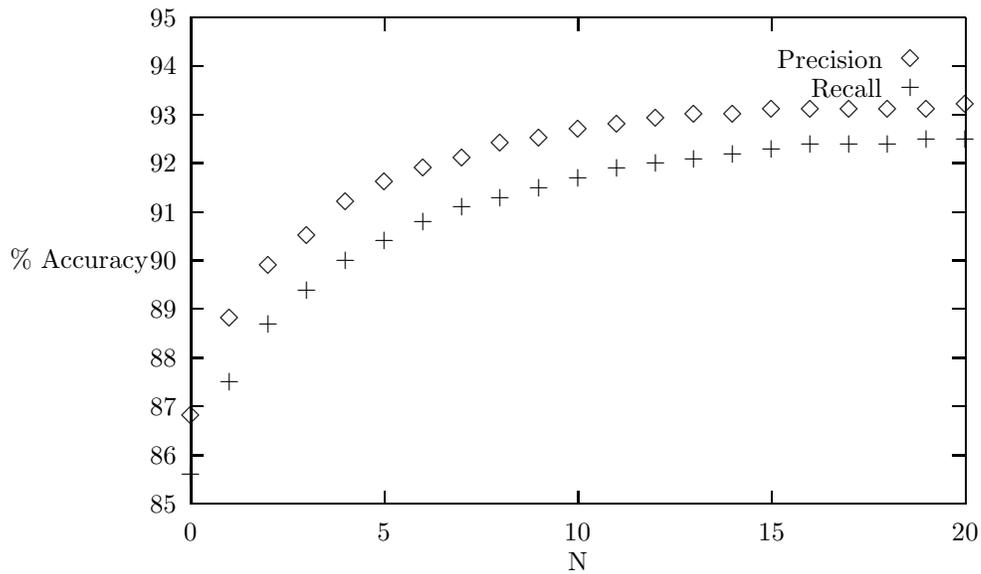
\begin{figure*}
\begin{center}
\setlength{\unitlength}{0.240900pt}
\ifx\plotpoint\undefined\newsavebox{\plotpoint}\fi
\sbox{\plotpoint}{\rule[-0.200pt]{0.400pt}{0.400pt}}%
\begin{picture}(1500,900)(0,0)
\font\gnuplot=cmr10 at 10pt
\gnuplot
\sbox{\plotpoint}{\rule[-0.200pt]{0.400pt}{0.400pt}}%
\put(220.0,113.0){\rule[-0.200pt]{0.400pt}{184.048pt}}
\put(220.0,113.0){\rule[-0.200pt]{4.818pt}{0.400pt}}
\put(198,113){\makebox(0,0)[r]{85}}
\put(1416.0,113.0){\rule[-0.200pt]{4.818pt}{0.400pt}}
\put(220.0,189.0){\rule[-0.200pt]{4.818pt}{0.400pt}}
\put(198,189){\makebox(0,0)[r]{86}}
\put(1416.0,189.0){\rule[-0.200pt]{4.818pt}{0.400pt}}
\put(220.0,266.0){\rule[-0.200pt]{4.818pt}{0.400pt}}
\put(198,266){\makebox(0,0)[r]{87}}
\put(1416.0,266.0){\rule[-0.200pt]{4.818pt}{0.400pt}}
\put(220.0,342.0){\rule[-0.200pt]{4.818pt}{0.400pt}}
\put(198,342){\makebox(0,0)[r]{88}}
\put(1416.0,342.0){\rule[-0.200pt]{4.818pt}{0.400pt}}
\put(220.0,419.0){\rule[-0.200pt]{4.818pt}{0.400pt}}
\put(198,419){\makebox(0,0)[r]{89}}
\put(1416.0,419.0){\rule[-0.200pt]{4.818pt}{0.400pt}}
\put(220.0,495.0){\rule[-0.200pt]{4.818pt}{0.400pt}}
\put(198,495){\makebox(0,0)[r]{90}}
\put(1416.0,495.0){\rule[-0.200pt]{4.818pt}{0.400pt}}
\put(220.0,571.0){\rule[-0.200pt]{4.818pt}{0.400pt}}
\put(198,571){\makebox(0,0)[r]{91}}
\put(1416.0,571.0){\rule[-0.200pt]{4.818pt}{0.400pt}}
\put(220.0,648.0){\rule[-0.200pt]{4.818pt}{0.400pt}}
\put(198,648){\makebox(0,0)[r]{92}}
\put(1416.0,648.0){\rule[-0.200pt]{4.818pt}{0.400pt}}
\put(220.0,724.0){\rule[-0.200pt]{4.818pt}{0.400pt}}
\put(198,724){\makebox(0,0)[r]{93}}
\put(1416.0,724.0){\rule[-0.200pt]{4.818pt}{0.400pt}}
\put(220.0,801.0){\rule[-0.200pt]{4.818pt}{0.400pt}}
\put(198,801){\makebox(0,0)[r]{94}}
\put(1416.0,801.0){\rule[-0.200pt]{4.818pt}{0.400pt}}
\put(220.0,877.0){\rule[-0.200pt]{4.818pt}{0.400pt}}
\put(198,877){\makebox(0,0)[r]{95}}
\put(1416.0,877.0){\rule[-0.200pt]{4.818pt}{0.400pt}}
\put(220.0,113.0){\rule[-0.200pt]{0.400pt}{4.818pt}}
\put(220,68){\makebox(0,0){0}}
\put(220.0,857.0){\rule[-0.200pt]{0.400pt}{4.818pt}}
\put(524.0,113.0){\rule[-0.200pt]{0.400pt}{4.818pt}}
\put(524,68){\makebox(0,0){5}}
\put(524.0,857.0){\rule[-0.200pt]{0.400pt}{4.818pt}}
\put(828.0,113.0){\rule[-0.200pt]{0.400pt}{4.818pt}}
\put(828,68){\makebox(0,0){10}}
\put(828.0,857.0){\rule[-0.200pt]{0.400pt}{4.818pt}}
\put(1132.0,113.0){\rule[-0.200pt]{0.400pt}{4.818pt}}
\put(1132,68){\makebox(0,0){15}}
\put(1132.0,857.0){\rule[-0.200pt]{0.400pt}{4.818pt}}
\put(1436.0,113.0){\rule[-0.200pt]{0.400pt}{4.818pt}}
\put(1436,68){\makebox(0,0){20}}
\put(1436.0,857.0){\rule[-0.200pt]{0.400pt}{4.818pt}}
\put(220.0,113.0){\rule[-0.200pt]{292.934pt}{0.400pt}}
\put(1436.0,113.0){\rule[-0.200pt]{0.400pt}{184.048pt}}
\put(220.0,877.0){\rule[-0.200pt]{292.934pt}{0.400pt}}
\put(45,495){\makebox(0,0){\% Accuracy}}
\put(828,23){\makebox(0,0){N}}
\put(220.0,113.0){\rule[-0.200pt]{0.400pt}{184.048pt}}
\put(1306,812){\makebox(0,0)[r]{Precision}}
\put(1350,812){\raisebox{-.8pt}{\makebox(0,0){$\Diamond$}}}
\put(220,251){\raisebox{-.8pt}{\makebox(0,0){$\Diamond$}}}
\put(281,403){\raisebox{-.8pt}{\makebox(0,0){$\Diamond$}}}
\put(342,487){\raisebox{-.8pt}{\makebox(0,0){$\Diamond$}}}
\put(402,533){\raisebox{-.8pt}{\makebox(0,0){$\Diamond$}}}
\put(463,587){\raisebox{-.8pt}{\makebox(0,0){$\Diamond$}}}
\put(524,617){\raisebox{-.8pt}{\makebox(0,0){$\Diamond$}}}
\put(585,640){\raisebox{-.8pt}{\makebox(0,0){$\Diamond$}}}
\put(646,655){\raisebox{-.8pt}{\makebox(0,0){$\Diamond$}}}
\put(706,678){\raisebox{-.8pt}{\makebox(0,0){$\Diamond$}}}
\put(767,686){\raisebox{-.8pt}{\makebox(0,0){$\Diamond$}}}
\put(828,701){\raisebox{-.8pt}{\makebox(0,0){$\Diamond$}}}
\put(889,709){\raisebox{-.8pt}{\makebox(0,0){$\Diamond$}}}
\put(950,717){\raisebox{-.8pt}{\makebox(0,0){$\Diamond$}}}
\put(1010,724){\raisebox{-.8pt}{\makebox(0,0){$\Diamond$}}}
\put(1071,724){\raisebox{-.8pt}{\makebox(0,0){$\Diamond$}}}
\put(1132,732){\raisebox{-.8pt}{\makebox(0,0){$\Diamond$}}}
\put(1193,732){\raisebox{-.8pt}{\makebox(0,0){$\Diamond$}}}
\put(1254,732){\raisebox{-.8pt}{\makebox(0,0){$\Diamond$}}}
\put(1314,732){\raisebox{-.8pt}{\makebox(0,0){$\Diamond$}}}
\put(1375,732){\raisebox{-.8pt}{\makebox(0,0){$\Diamond$}}}
\put(1436,739){\raisebox{-.8pt}{\makebox(0,0){$\Diamond$}}}
\put(1306,767){\makebox(0,0)[r]{Recall}}
\put(1350,767){\makebox(0,0){$+$}}
\put(220,159){\makebox(0,0){$+$}}
\put(281,304){\makebox(0,0){$+$}}
\put(342,396){\makebox(0,0){$+$}}
\put(402,449){\makebox(0,0){$+$}}
\put(463,495){\makebox(0,0){$+$}}
\put(524,526){\makebox(0,0){$+$}}
\put(585,556){\makebox(0,0){$+$}}
\put(646,579){\makebox(0,0){$+$}}
\put(706,594){\makebox(0,0){$+$}}
\put(767,610){\makebox(0,0){$+$}}
\put(828,625){\makebox(0,0){$+$}}
\put(889,640){\makebox(0,0){$+$}}
\put(950,648){\makebox(0,0){$+$}}
\put(1010,655){\makebox(0,0){$+$}}
\put(1071,663){\makebox(0,0){$+$}}
\put(1132,671){\makebox(0,0){$+$}}
\put(1193,678){\makebox(0,0){$+$}}
\put(1254,678){\makebox(0,0){$+$}}
\put(1314,678){\makebox(0,0){$+$}}
\put(1375,686){\makebox(0,0){$+$}}
\put(1436,686){\makebox(0,0){$+$}}
\end{picture}
\caption{Precision \& recall of a ``perfect'' reranking scheme for the top $N$
parses of section 23 of the WSJ Treebank, as a function of $N$. Evaluation ignores quotation marks.}
\label{topnpr}
\end{center}
\end{figure*}

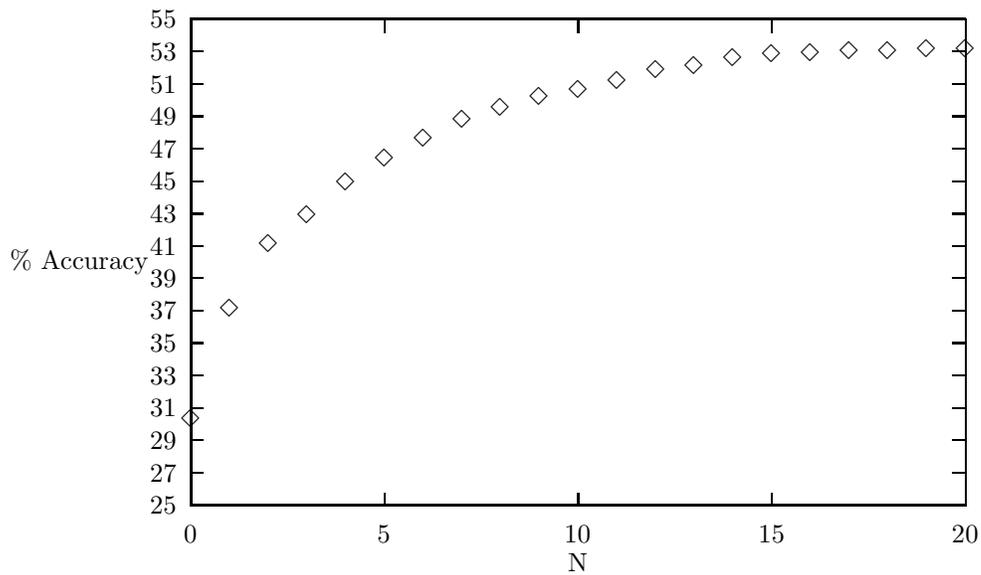
\begin{figure*}
\begin{center}
\setlength{\unitlength}{0.240900pt}
\ifx\plotpoint\undefined\newsavebox{\plotpoint}\fi
\begin{picture}(1500,900)(0,0)
\font\gnuplot=cmr10 at 10pt
\gnuplot
\sbox{\plotpoint}{\rule[-0.200pt]{0.400pt}{0.400pt}}%
\put(220.0,113.0){\rule[-0.200pt]{0.400pt}{184.048pt}}
\put(220.0,113.0){\rule[-0.200pt]{4.818pt}{0.400pt}}
\put(198,113){\makebox(0,0)[r]{25}}
\put(1416.0,113.0){\rule[-0.200pt]{4.818pt}{0.400pt}}
\put(220.0,164.0){\rule[-0.200pt]{4.818pt}{0.400pt}}
\put(198,164){\makebox(0,0)[r]{27}}
\put(1416.0,164.0){\rule[-0.200pt]{4.818pt}{0.400pt}}
\put(220.0,215.0){\rule[-0.200pt]{4.818pt}{0.400pt}}
\put(198,215){\makebox(0,0)[r]{29}}
\put(1416.0,215.0){\rule[-0.200pt]{4.818pt}{0.400pt}}
\put(220.0,266.0){\rule[-0.200pt]{4.818pt}{0.400pt}}
\put(198,266){\makebox(0,0)[r]{31}}
\put(1416.0,266.0){\rule[-0.200pt]{4.818pt}{0.400pt}}
\put(220.0,317.0){\rule[-0.200pt]{4.818pt}{0.400pt}}
\put(198,317){\makebox(0,0)[r]{33}}
\put(1416.0,317.0){\rule[-0.200pt]{4.818pt}{0.400pt}}
\put(220.0,368.0){\rule[-0.200pt]{4.818pt}{0.400pt}}
\put(198,368){\makebox(0,0)[r]{35}}
\put(1416.0,368.0){\rule[-0.200pt]{4.818pt}{0.400pt}}
\put(220.0,419.0){\rule[-0.200pt]{4.818pt}{0.400pt}}
\put(198,419){\makebox(0,0)[r]{37}}
\put(1416.0,419.0){\rule[-0.200pt]{4.818pt}{0.400pt}}
\put(220.0,470.0){\rule[-0.200pt]{4.818pt}{0.400pt}}
\put(198,470){\makebox(0,0)[r]{39}}
\put(1416.0,470.0){\rule[-0.200pt]{4.818pt}{0.400pt}}
\put(220.0,520.0){\rule[-0.200pt]{4.818pt}{0.400pt}}
\put(198,520){\makebox(0,0)[r]{41}}
\put(1416.0,520.0){\rule[-0.200pt]{4.818pt}{0.400pt}}
\put(220.0,571.0){\rule[-0.200pt]{4.818pt}{0.400pt}}
\put(198,571){\makebox(0,0)[r]{43}}
\put(1416.0,571.0){\rule[-0.200pt]{4.818pt}{0.400pt}}
\put(220.0,622.0){\rule[-0.200pt]{4.818pt}{0.400pt}}
\put(198,622){\makebox(0,0)[r]{45}}
\put(1416.0,622.0){\rule[-0.200pt]{4.818pt}{0.400pt}}
\put(220.0,673.0){\rule[-0.200pt]{4.818pt}{0.400pt}}
\put(198,673){\makebox(0,0)[r]{47}}
\put(1416.0,673.0){\rule[-0.200pt]{4.818pt}{0.400pt}}
\put(220.0,724.0){\rule[-0.200pt]{4.818pt}{0.400pt}}
\put(198,724){\makebox(0,0)[r]{49}}
\put(1416.0,724.0){\rule[-0.200pt]{4.818pt}{0.400pt}}
\put(220.0,775.0){\rule[-0.200pt]{4.818pt}{0.400pt}}
\put(198,775){\makebox(0,0)[r]{51}}
\put(1416.0,775.0){\rule[-0.200pt]{4.818pt}{0.400pt}}
\put(220.0,826.0){\rule[-0.200pt]{4.818pt}{0.400pt}}
\put(198,826){\makebox(0,0)[r]{53}}
\put(1416.0,826.0){\rule[-0.200pt]{4.818pt}{0.400pt}}
\put(220.0,877.0){\rule[-0.200pt]{4.818pt}{0.400pt}}
\put(198,877){\makebox(0,0)[r]{55}}
\put(1416.0,877.0){\rule[-0.200pt]{4.818pt}{0.400pt}}
\put(220.0,113.0){\rule[-0.200pt]{0.400pt}{4.818pt}}
\put(220,68){\makebox(0,0){0}}
\put(220.0,857.0){\rule[-0.200pt]{0.400pt}{4.818pt}}
\put(524.0,113.0){\rule[-0.200pt]{0.400pt}{4.818pt}}
\put(524,68){\makebox(0,0){5}}
\put(524.0,857.0){\rule[-0.200pt]{0.400pt}{4.818pt}}
\put(828.0,113.0){\rule[-0.200pt]{0.400pt}{4.818pt}}
\put(828,68){\makebox(0,0){10}}
\put(828.0,857.0){\rule[-0.200pt]{0.400pt}{4.818pt}}
\put(1132.0,113.0){\rule[-0.200pt]{0.400pt}{4.818pt}}
\put(1132,68){\makebox(0,0){15}}
\put(1132.0,857.0){\rule[-0.200pt]{0.400pt}{4.818pt}}
\put(1436.0,113.0){\rule[-0.200pt]{0.400pt}{4.818pt}}
\put(1436,68){\makebox(0,0){20}}
\put(1436.0,857.0){\rule[-0.200pt]{0.400pt}{4.818pt}}
\put(220.0,113.0){\rule[-0.200pt]{292.934pt}{0.400pt}}
\put(1436.0,113.0){\rule[-0.200pt]{0.400pt}{184.048pt}}
\put(220.0,877.0){\rule[-0.200pt]{292.934pt}{0.400pt}}
\put(45,495){\makebox(0,0){\% Accuracy}}
\put(828,23){\makebox(0,0){N}}
\put(220.0,113.0){\rule[-0.200pt]{0.400pt}{184.048pt}}
\put(220,248){\raisebox{-.8pt}{\makebox(0,0){$\Diamond$}}}
\put(281,421){\raisebox{-.8pt}{\makebox(0,0){$\Diamond$}}}
\put(342,523){\raisebox{-.8pt}{\makebox(0,0){$\Diamond$}}}
\put(402,569){\raisebox{-.8pt}{\makebox(0,0){$\Diamond$}}}
\put(463,620){\raisebox{-.8pt}{\makebox(0,0){$\Diamond$}}}
\put(524,658){\raisebox{-.8pt}{\makebox(0,0){$\Diamond$}}}
\put(585,689){\raisebox{-.8pt}{\makebox(0,0){$\Diamond$}}}
\put(646,719){\raisebox{-.8pt}{\makebox(0,0){$\Diamond$}}}
\put(706,737){\raisebox{-.8pt}{\makebox(0,0){$\Diamond$}}}
\put(767,755){\raisebox{-.8pt}{\makebox(0,0){$\Diamond$}}}
\put(828,765){\raisebox{-.8pt}{\makebox(0,0){$\Diamond$}}}
\put(889,780){\raisebox{-.8pt}{\makebox(0,0){$\Diamond$}}}
\put(950,796){\raisebox{-.8pt}{\makebox(0,0){$\Diamond$}}}
\put(1010,803){\raisebox{-.8pt}{\makebox(0,0){$\Diamond$}}}
\put(1071,816){\raisebox{-.8pt}{\makebox(0,0){$\Diamond$}}}
\put(1132,821){\raisebox{-.8pt}{\makebox(0,0){$\Diamond$}}}
\put(1193,824){\raisebox{-.8pt}{\makebox(0,0){$\Diamond$}}}
\put(1254,826){\raisebox{-.8pt}{\makebox(0,0){$\Diamond$}}}
\put(1314,826){\raisebox{-.8pt}{\makebox(0,0){$\Diamond$}}}
\put(1375,829){\raisebox{-.8pt}{\makebox(0,0){$\Diamond$}}}
\put(1436,829){\raisebox{-.8pt}{\makebox(0,0){$\Diamond$}}}
\end{picture}
\caption{Exact match of a ``perfect'' reranking scheme for the top $N$ 
parses of section 23 of the WSJ Treebank, as a function of $N$. Evaluation ignores quotation marks.}
\label{topnx}
\end{center}
\end{figure*}

It is often advantageous to produce the top $N$ parses instead of 
just the top 1, since additional information can be used in a secondary 
model that re-orders the top $N$ and
hopefully improves the quality of the top ranked parse.
Suppose there exists a ``perfect'' reranking scheme that,
for each sentence, magically picks the {\em best} parse from the top $N$ parses
produced by the maximum entropy parser, where
the {\em best} parse has the highest average precision and recall
when compared to the treebank parse.
The performance of this ``perfect'' scheme is then an upper bound
on the performance of any reranking scheme that might be used to reorder
the top $N$ parses. 
Figure~\ref{topnpr} shows that the ``perfect'' scheme would achieve roughly 93\%
precision and recall, which is a dramatic increase over the top 1 accuracy
of 87\% precision and 86\% recall.
Figure~\ref{topnx} shows that the ``Exact Match'', which 
counts the percentage of times the proposed parse $P$ is identical (excluding
POS tags) to the treebank parse $T$, rises substantially to about 53\% from
30\% when the ``perfect'' scheme is applied.
For this reason, research into reranking schemes appears to be a promising
step towards the goal of improving parsing accuracy.

\begin{table}
\begin{center}
\begin{tabular}{|c|r|r|r|r|r|}\hline
Parser&Precision&Recall\\ \hline
Maximum Entropy$^\diamond$&86.8\%&85.6\%\\ \hline
Maximum Entropy$^\star$&87.5\%&86.3\%\\ \hline
\cite{collins:acl96}$^\star$&85.7\%&85.3\%\\ \hline
\cite{magerman:acl95}$^\star$&84.3\%&84.0\%\\ \hline
\end{tabular}
\end{center}
\caption{Results on 2416 sentences of section 23 (0 to 100 words in length) of the WSJ Treebank.
Evaluations marked with $^\diamond$ ignore quotation marks.
Evaluations marked with $^\star$ collapse the distinction between {\tt ADVP} and {\tt PRT}, and
ignore {\em all} punctuation.}
\label{results}
\end{table}

\section{Comparison With Previous Work}
\label{discussion}

The two parsers which have previously reported 
the best accuracies on the Penn Treebank Wall St.~Journal
are the bigram parser described in \cite{collins:acl96} and
the SPATTER parser described in \cite{ibm:spatter,magerman:acl95}.
The parser presented here outperforms both the bigram parser and
the SPATTER parser, and uses different modelling technology and
different information to drive its decisions.

The bigram parser is a statistical CKY-style chart parser,
which uses cooccurrence statistics of head-modifier pairs to find 
the best parse. The maximum entropy parser is a
statistical shift-reduce style parser that 
cannot always access head-modifier pairs.
For example, the \feat{checkcons}{m,n} predicate of the maximum entropy parser
may use two words such that {\em neither} is the intended
head of the proposed consituent that the \proc{check} procedure must judge.
And unlike the bigram parser, the maximum entropy parser cannot use 
head word information besides ``flat'' chunks in the right context.

The bigram parser uses a backed-off estimation scheme that
is customized for a particular task, whereas the maximum entropy
parser uses a general purpose modelling technique.
This allows the maximum entropy parser to easily integrate
varying kinds of features, such as those for punctuation, 
whereas the bigram parser uses hand-crafted punctuation rules.
Furthermore, the customized estimation framework of the bigram parser 
must use information that has been carefully selected for its value,
whereas the maximum entropy framework robustly
integrates any kind of information, obviating the need to
screen it first.

The SPATTER parser is a history-based parser that
uses decision tree models to guide the operations of
a few tree building procedures. It differs from
the maximum entropy parser in how it builds trees and
more critically, in how its decision trees use information.
The SPATTER decision trees use predicates on word classes 
created with a statistical clustering technique,
whereas the maximum entropy parser uses predicates
that contain merely the words themselves, and thus lacks
the need for a (typically expensive) word clustering procedure.
Furthermore, the top $K$ BFS search heuristic appears to 
be much simpler than the stack decoder algorithm outlined
in \cite{magerman:acl95}. 

\section{Conclusion}

The maximum entropy parser presented here achieves
a parsing accuracy which exceeds the best previously published results,
and parses a test sentence in linear observed time, with respect to the sentence length.
It uses simple and concisely specified predicates which can added or modified quickly
with little human effort under the maximum entropy framework. 
Lastly, this paper clearly demonstrates that schemes for reranking the top 20
parses deserve research effort since they could yield vastly better accuracy results.

\section{Acknowledgements}

Many thanks to Mike Collins and Professor Mitch Marcus from
the University of Pennsylvania for their helpful comments on this work.

\bibliography{refs}

\end{document}